\documentclass[lettersize,journal]{IEEEtran}
\usepackage{amsmath,amssymb, amsfonts}
\usepackage{algorithmic}
\usepackage{algorithm}
\usepackage{array}
\usepackage{color} 
\usepackage[caption=false,font=normalsize,labelfont=sf,textfont=sf]{subfig}
\usepackage{textcomp}
\usepackage{stfloats}
\usepackage{url}
\usepackage{verbatim}
\usepackage{graphicx}
\usepackage{cite}
\usepackage{multirow}
\usepackage{float}
\begin{document}
\title{INFWIDE: Image and Feature Space Wiener Deconvolution Network for Non-blind Image Deblurring in Low-Light Conditions}

\author{Zhihong~Zhang,
        Yuxiao~Cheng,
        Jinli~Suo,
        Liheng~Bian,
        and~Qionghai~Dai,~\IEEEmembership{Senior~Member,~IEEE}%
\thanks{\noindent All the authors except for Liheng Bian are with the Department of Automation, and the Institute for Brain and Cognitive Sciences, Tsinghua University, Beijing 100084, China.
E-mails: zhangzh19@mails.tsinghua.edu.cn, chengyx18@mails.tsinghua.edu.cn, jlsuo@tsinghua.edu.cn, qhdai@tsinghua.edu.cn.}
\thanks{Liheng Bian is with Beijing National Research Center for Information Science and Technology, and Advanced Research Institute of Multidisciplinary Science, Beijing Institute of Technology, Beijing 100081, China. 
E-mail: bian@bit.edu.cn}
\thanks{{Corresponding author: Jinli Suo}}
}




\maketitle

\begin{abstract}
Under low-light environment, handheld photography suffers from severe camera shake under long exposure settings. Although existing deblurring algorithms have shown promising performance on well-exposed blurry images, they still cannot cope with low-light snapshots. 
Sophisticated noise and saturation regions are two dominating challenges in practical low-light deblurring: the former violates the Gaussian or Poisson assumption widely used in most existing algorithms and thus degrades their performance badly, while the latter introduces non-linearity to the classical convolution-based blurring model and makes the deblurring task even challenging. 
In this work, we propose a novel non-blind deblurring method dubbed image and feature space Wiener deconvolution network (INFWIDE) to tackle these problems systematically.
In terms of algorithm design, INFWIDE proposes a two-branch architecture, which explicitly removes noise and hallucinates saturated regions in the image space and suppresses ringing artifacts in the feature space, and integrates the two complementary outputs with a subtle multi-scale fusion network for high quality night photograph deblurring.
For effective network training, we design a set of loss functions integrating a forward imaging model and backward reconstruction to form a close-loop regularization to secure good convergence of the deep neural network.
Further, to optimize INFWIDE's applicability in real low-light conditions, a physical-process-based low-light noise model is employed to synthesize realistic noisy night photographs for model training. 
Taking advantage of the traditional Wiener deconvolution algorithm's physically driven characteristics and deep neural network's representation ability, INFWIDE can recover fine details while suppressing the unpleasant artifacts during deblurring. 
Extensive experiments on synthetic data and real data demonstrate the superior performance of the proposed approach.
\end{abstract}


\section{Introduction}
\begin{figure}[t]
  \centering
  \includegraphics[width=\linewidth]{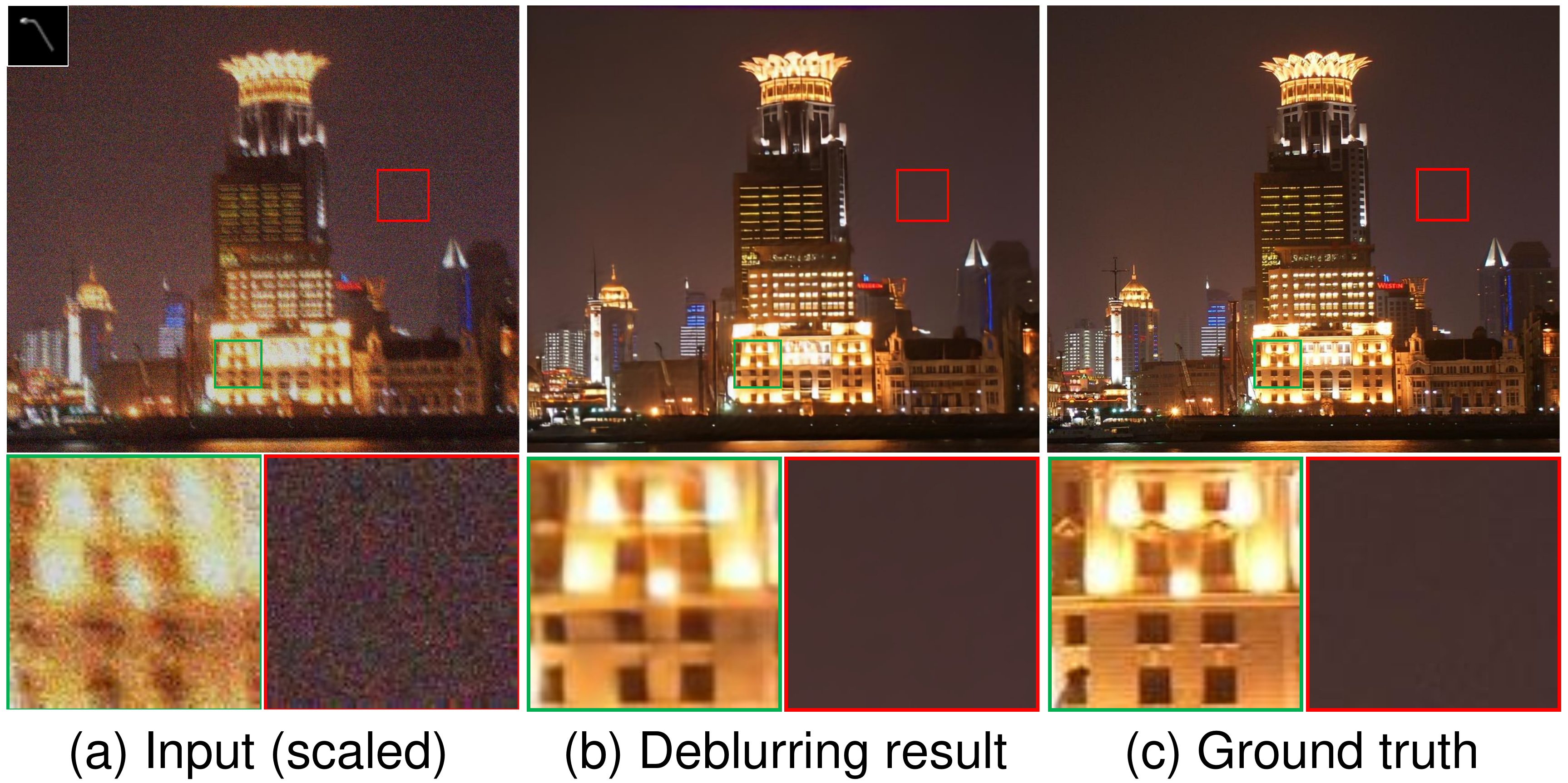}
  \caption{An example of non-blind low-light deblurring with the proposed method (INFWIDE). (a) A typical low-light blurry image degenerated by a kernel in the topleft inset, with saturation in highlight regions (green box) and severe noise in dark regions (red box). We linearly scale the image for better visualization. (b) The deblurring result of INFWIDE. (c) Ground truth.}
  \label{fig:teaser}
\end{figure}

\IEEEPARstart{I}{mage} deblurring aims to recover sharp images from their blurry counterparts, and non-blind image deblurring further assumes that the blur kernel is known or could be calibrated beforehand. Considering that blur is a widespread degeneration which can be caused by system defect, object motion, limited depth-of-field, camera shake, etc., and will greatly degrade the sharpness, deblurring has become an indispensable post-processing step in various applications including photography \cite{su2017DeepVideo}, microscopy \cite{yanny2022DeepLearning},  telemetry\cite{ma2009DeblurringHighly}, astronomy \cite{liu2019BlindDeblurring}, etc. Generally, the blurring process can be mathematically formulated as
\begin{equation}
\label{eq: blur_model}
\mathbf{y} = \mathbf{x} \ast \mathbf{k} + \mathbf{n},
\end{equation}
where $\mathbf{y}$, $\mathbf{x}$, $\mathbf{k}$ and $\mathbf{n}$ denote the captured noisy blurry image, latent clean sharp image and additive noise, respectively; $\ast$ represents the convolution operation. From this blurring model, we can figure out that the deblurring problem is a highly ill-posed inverse problem whose complexity is relevant to the blur kernel $\mathbf{k}$ and noise $\mathbf{n}$. In low-light conditions, a longer exposure is usually required to collect enough photons onto the sensor, which will thus result in a more complex blur kernel. Even under long exposure settings, the dark regions still suffer from noise, as exemplified by the red highlighted region in Fig.~\ref{fig:teaser}(a). Besides, noise in low-light conditions follows a more sophisticated model that cannot be simplified to be a Gaussian or Poisson distribution as most deblurring algorithms do. These problems make low-light deblurring remain a big challenge. 


Apart from the aforementioned problems, nighttime blurry images also feature high dynamic range (HDR) and dazzle light against the dark background, especially in cityscape night photography, as shown in the green highlighted regions in  Fig.~\ref{fig:teaser}(a). Considering that the pixels within the bright regions will generally cause overexposure and be clipped to the sensor's saturation value, Eq.~\eqref{eq: blur_model} can be further adapted to
\begin{equation}
\label{eq: night_blur_model}
\mathbf{y} = \mathrm{Clip}(\mathbf{x} \ast \mathbf{k} + \mathbf{n}),
\end{equation}
where $\mathrm{Clip}(\cdot)$ denotes the non-linear clipping function (i.e. $\min(\cdot,1)$)\footnote{We assume that the gray values are normalized to [0, 1], similarly hereinafter.}.
Saturated pixels violate the common assumption that the image blurring model is linear and often bring about ringing artifacts in deblurred images. Therefore, how to deal with these saturated pixels is also a hard problem to solve in low-light deblurring.

Conventional non-blind deblurring algorithms can be roughly divided into two categories as reconstructing sharp images in either frequency-domain or spatial-domain. Early classical frequency-domain algorithms like inverse filtering \cite{sankhe2011DeblurringGrayscale} and Wiener filtering \cite{wiener1949ExtrapolationInterpolation} transform the non-blind image deblurring problem from image space to the Fourier space, and have lower computation complexity with the aid of fast Fourier transform (FFT) \cite{nussbaumer1981FastFourier}. However, this class of algorithms is based on the linear blurring model formulated in Eq.~\eqref{eq: blur_model} and hardly explores the underlying image priors, which limits their deblurring performance to a large extent. Besides, their performance is also greatly affected by noise and saturation, which will result in severe ringing artifacts in deblurred images. 

Spatial-domain approaches generally formulate the non-blind image deblurring as a minimization problem of the following form
\begin{equation}
\label{map}
 \mathbf{\hat{x}} =  \mathop{\arg\min}\limits_{\mathbf{x}} \mathcal{D}(\mathbf{x} \ast \mathbf{k}, \mathbf{y})+\alpha\mathcal{R}(\mathbf{x}) 
\end{equation}
where $\mathcal{D}$ and $\mathcal{R}$ represent the data-fidelity term and regularization term, respectively; the scalar weight $\alpha$ is used to balance these two terms in optimization.
From the probabilistic perspective, different data-fidelity terms such as $l_2$ norm and $l_1$ norm can be derived from the negative log-likelihood with the assumption of corresponding noise distribution \cite{whyte2014DeblurringShaken}. In respect of the regularization term, various image prior-based penalties such as sparsity \cite{xu2013UnnaturalL0}, total variation \cite{wang2008NewAlternating} and hyper-Laplacian \cite{krishnan2009FastImage} have been exploited to guarantee the solution's convergence towards desired clear sharp images. With the constraint of model-based image priors, these spatial-domain methods have achieved a great progress compared with traditional frequency-domain algorithms. However. it is still challenging for them to deal with severe noise in low-light conditions. Besides,  they can hardly prevent ringing artifacts in presence of saturation neither.

In recent decades, learning-based algorithms have become the dominant image deblurring approaches. From early ``shallow-learning" methods like Gaussian mixture model (GMM) \cite{joshi2009ImageDeblurring, zoran2011LearningModels} and dictionary learning \cite{hu2010SingleImage, ma2013DictionaryLearning} to recent deep neural networks (DNNs) \cite{son2017FastNonblind,zhang2017LearningFully,ren2018DeepNonBlind,vasu2018NonblindDeblurring, chen2021LearningNonBlind,koh2021SingleimageDeblurring}, learning-based deblurring methods have made remarkable progress by digging deeper into data-driven image priors. Apart from pure DNN approaches, combining conventional filtering-based or optimization-based methods with recent learning-based DNNs to design novel physical-guided data-driven algorithms has also become a new research highlight in this field\cite{corbineau2019LearnedImage,li2020EfficientInterpretable,dong2020DeepWiener,sanghvi2021PhotonLimitedDeblurring}. 

In this paper, we propose a novel non-blind image deblurring neural network named INFWIDE to tackle the sophisticated noise and saturation issue in low-light conditions. 
INFWIDE leverages conventional physically-driven Wiener deconvolution and data-driven deep neural networks to take advantageous strengths of both regimes. 
Such a scheme can be implemented in either image space or feature space. Generally, image space processing is ready for extensions such as incorporating noise suppressor or compensating clipped values explicitly. 
However, the deblurring results of image space processing 
are prone to over-smoothness or ringing artifacts. On the contrary, 
recent works \cite{dong2020DeepWiener, dong2021DWDNDeep} have demonstrated that feature space Wiener deconvolution has an advantage over the image space counterpart in terms of recovering fine details and suppressing ringing artifacts in well-exposed scenarios, while its limitation lies in the failure of dealing with large saturated regions and heavy noise.
Therefore, an intuitive but effective way is to take advantage of both image space and feature space deconvolution results and fuse them together to boost the final deblurring performance. Bearing this in mind, we design a multi-scale cross residual fusion module to exploit the complementary information in both spaces and fusion them across multi-levels to generate a clear sharp image with fine details. 

Under the above two-branch architecture, we propose three strategies to address the challenges of long exposure sharp imaging in low light conditions systematically. 
Firstly, to mitigate the influence of noise and saturation, INFWIDE explicitly estimates noise level and is equipped with an enhancement module in the image branch to recover a noise-free non-clipped blurry image obeying the linear blurring model, which facilitates the subsequent Wiener deconvolution. 
Secondly, to advance network convergence, we further exploit the analytical blurring model and introduce a physics-constrained `reblurring loss' during training, which reblurs the deblurring result and minimizes its gap with the ground-truth clear non-clipped blurry image. With `reblurring loss', we can efficiently build a close-loop regularization to guarantee the network's convergence within a feasible region.
Moreover, a large nighttime image dataset and a comprehensive physical-process-based noise model are also employed to synthesize realistic low-light training data. In this manner, the proposed method can attain a good generalization ability in practical applications.
Benefiting from these techniques, INFWIDE can recover fine details buried in blurry snapshots while preventing unpleasant artifacts, as shown in Fig.~\ref{fig:teaser}(b).

In a nutshell, the main contributions of this work can be summarized as follows.
\begin{itemize}
    \item We propose a novel non-blind deblurring network named INFWIDE to deal with the sophisticated noise and saturation problem in low-light conditions by incorporating model-driven Wiener deconvolution and data-driven regularization in spatial and frequency spaces. 
    \item We design a loss function integrating the forward imaging model and backward reconstruction to form a closed-loop regulation to guide the network's convergence towards a reasonable direction. 
    \item We collect a large nighttime image dataset and employ a physical-process-based noise model to synthesize realistic low-light blurry images to train INFWIDE, which empowers it with good generalization ability in practical applications.
    \item We qualitatively and quantitatively evaluate INFWIDE with simulated and real data experiments. The results demonstrate INFWIDE's superior performance to existing methods.
\end{itemize}

The rest of this paper is organized as follows: Firstly, We briefly review existing works for low-light image deblurring in Section \ref{sec:related_work}. Then Section \ref{sec:our_method} introduces the architecture of the proposed INFWIDE along with loss function settings, and Section \ref{sec:data_gen} presents the procedures for low-light image dataset collection and realistic low-light noise and saturation synthesis. In Section \ref{sec:exp}, we describe the method implementation details and provide comprehensive experiments to demonstrate the performance of the proposed approach. Lastly, the paper concludes in Section \ref{sec:convlusion} with summary and some discussions.

\section{Related Work}
\label{sec:related_work}
Non-blind image deblurring is a classical image restoration problem which has been studied for decades \cite{yuan2008ProgressiveInterscale,cho2011HandlingOutliers,zhang2017LearningFully,anger2018ModelingRealistic,vasu2018NonblindDeblurring,eboli2020EndtoendInterpretable}. Existing works on this topic are mainly based on the linear blurring model of Eq.~\eqref{eq: blur_model} and seek to improving the deblurring performance under normal light conditions.  However, as mentioned above, blurring is more commonly seen in low-light conditions, where longer exposure is required, and complex noise and saturation problems in low-light environment exert more challenges on image deblurring. Therefore, as a meaningful branch of the deblurring task, low-light deblurring is actually still an open challenge. In this section, we review current status of low-light deblurring from the aforementioned two aspects, i.e., handling the sophisticated noise and tackling the saturation problem, and give a brief summation afterwards.

\subsection{Handling Sophisticated Noise}
\label{sec:related_work_noise}
Noise is an inevitable and undesirable factor in digital photography, and generally has a significant influence on blur kernel estimation and non-blind deblurring algorithms \cite{anger2019EfficientBlind, dasgupta2022ComparativeAnalysis}. Existing works on non-blind image deblurring often simply assume that the noise obeys Gaussian distribution \cite{son2017FastNonblind, vasu2018NonblindDeblurring, corbineau2019LearnedImage, miao2020HandlingNoise}, which may achieve relatively ideal performance under most well-illuminated scenes. Nevertheless, noise model in low-light conditions is usually more sophisticated, and the mismatch of noise distribution between deblurring algorithms and real blurry images will degrade the deblurring performance badly.

An intuitive approach for improving deblurring algorithms' performance in low-light conditions is to design an elaborate noise model, which fits photon-limited images well. Considering that photon shot noise plays a dominant role in low-light conditions, many low-light deblurring algorithms model the noise with Poisson distribution instead of widely used Gaussian distribution to adapt to real-captured blurry images. For example,
in \cite{ma2013DictionaryLearning}, Ma et al. proposed the first Poisson image deblurring algorithm via a patch-based sparse representation prior (i.e. a learned dictionary) to handle blurry images corrupted by Poisson noise. 
In \cite{chowdhury2020NonblindBlind}, a fractional-order total variation regularization was proposed to remove the blur and Poisson noise simultaneously, and restore latent images with high-order smoothness.
Most recently, in \cite{sanghvi2021PhotonLimited}, Sanghvi et al. formulated the non-blind deblurring of photon-limited blurry images as a Poisson linear inverse problem, and proposed an end-to-end unrolling network to tackle this problem by using a three-operator splitting technique to turn all sub-routines differentiable. However, it is worth noting that Poisson noise is not the only noise type contained in photon-starved images, and thus aforementioned approaches are only applicable to limited scenarios with extremely low-light illumination in practical applications.

On the other hand, to mitigate severe noise's influence on low-light deblurring, some algorithms incorporate denoising into deblurring to boost the overall image restoration performance. For instance, in \cite{anger2019EfficientBlind}, Anger et al. showed that the $l_0$ gradient prior-based blur kernel estimation methods can be adapted to handle severe noise efficiently. Besides, they also demonstrated that significant improvement can be attained for fast non-blind deconvolution by firstly denoising the blurry image. In \cite{miao2020HandlingNoise}, a cascaded framework composed of a denoising sub-network and a deblurring sub-network was proposed. By jointly training these two sub-networks, the effect of residual noise could be reduced, which thus empowered the method with more robustness under heavy noise.

\subsection{Tackling Saturation Problem}
Apart from sophisticated noise, night photography also features saturation problem due to high dynamic range. Saturated pixels violate linear blurring model, and will bring about severe ringing artifacts if not handled well \cite{whyte2014DeblurringShaken}.

To alleviate saturated pixels' influence on overall deblurring performance, a direct way is to simply discard these outlier pixels and only process the rest of the image which obeys the linear convolution model shown in Eq.~\eqref{eq: blur_model}. For example, Harmeling et al. estimated the saturation mask directly by applying a close-to-1 thresholding to the blurred image, i.e., pixels with intensity above this threshold were treated as saturated pixels and excluded in subsequent deblurring process\cite{harmeling2010MultiframeBlind}. Cho et al. adopted an Expectation-Maximization-based framework for non-blind deblurring, and adaptively calculated and refreshed the saturation mask by blurring the current estimation of the sharp image and thresholding it in each deblurring iteration \cite{cho2011HandlingOutliers}. A similar strategy was adopted by Hu et al. in \cite{hu2014DeblurringLowlight}. Recently, a deep convolutional neural network (CNN) based approach was also proposed by Dong et al. to directly estimate the confidence map of saturated pixels and other outliers to facilitate the following deblurring process \cite{dong2021DeepOutlier}.

There are also some other approaches that develop specially-designed data-fidelity terms to fit the non-linear blurring model shown in Eq.~\eqref{eq: night_blur_model}. Specifically, Whyte et al. introduced a non-linear response function to model the effect of saturation, and involved its smooth, continuously differentiable approximation into classical Richardson-Lucy algorithm to prevent ringing artifacts in bright regions\cite{whyte2014DeblurringShaken}. In \cite{pan2016RobustKernel}, Pan et al. proposed a robust energy function to describe the property of outliers, and incorporated it into the maximum a posteriori (MAP) framework to facilitate the latent sharp image restoration \cite{pan2016RobustKernel}. Most recently, a data-driven non-blind deblurring network (NBDN) that learns both the fidelity and prior terms was developed by Chen et al. in \cite{chen2021LearningNonBlind}. Benefiting from the superior representation ability of DNN, NBDN achieved better performance in low-light deblurring compared with traditional optimization-based methods.

\begin{figure*}[!t]
  \centering
  \includegraphics[width=\linewidth]{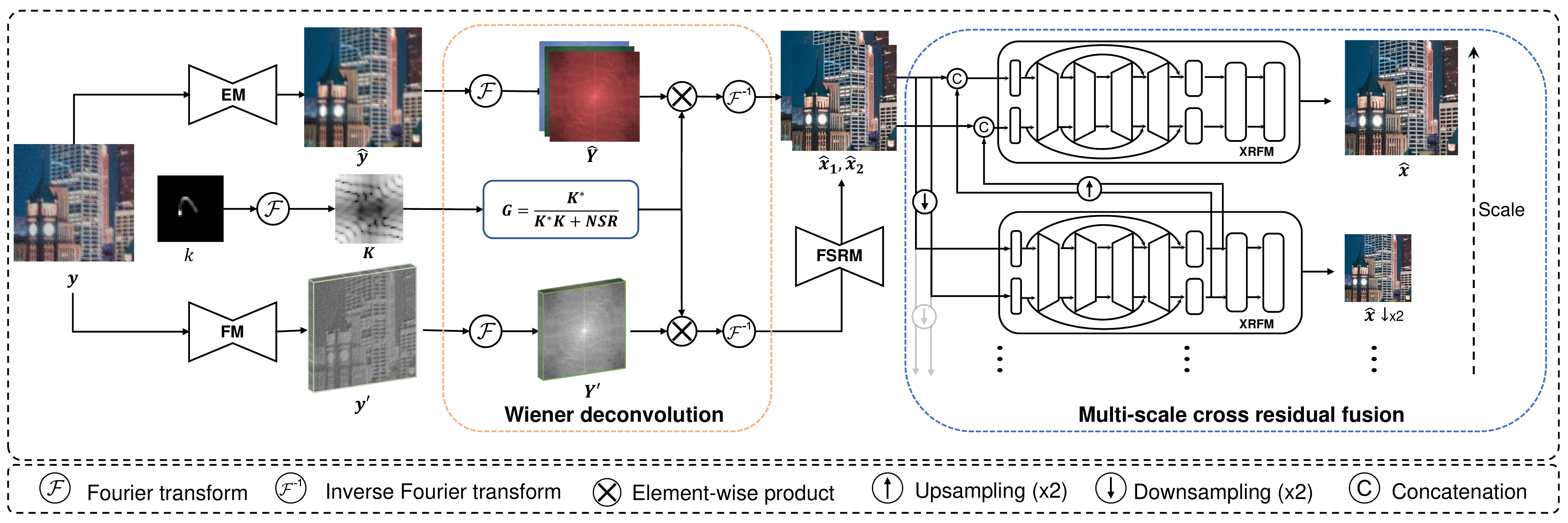}
  \caption{The architecture of INFWIDE. \textbf{Image space}: a clear and non-clipped image $\hat{\mathbf{y}}$ is first estimated from the low-light blurry image $\mathbf{y}$ by the enhancement module (EM). Then an intermediate deblurring result $\hat{\mathbf{x}}_1$ is generated by image space Wiener deconvolution. \textbf{Feature space}: a set of feature maps $\mathbf{y'}$ is first extracted from $\mathbf{y}$ by the feature module (FM). Then another intermediate deblurring result $\hat{\mathbf{x}}_2$ is attained by feature space Wiener deconvolution and feature space refine (FSRM).  \textbf{Fusion}: $\hat{\mathbf{x}}_1$ and $\hat{\mathbf{x}}_2$ are fused by a multi-sale cross residual fusion module (XRFM) to form the final deblurring result $\hat{\mathbf{x}}$.}
  \label{fig:net_infwide}
\end{figure*}

\subsection{Summary and relation with our work}
Blur, noise, and saturation are three major challenges for night photography. Although there exist plenty of early attempts 
to address the noise or saturation problem, few works take full consideration of both factors, and their performance is also limited in real applications. On the one hand, low-light images suffer from more sophisticated noise, which cannot be simply modeled as Gaussian or Poisson distribution. The mismatch between employed and real noise model will result in severe artifacts. On the other hand, although the frequently-used ``mask and exclude" strategy can efficiently prevent ringing artifacts caused by saturated pixels, the cost is to sacrifice useful information around saturation regions. With recent advances in deep learning, some learning-based algorithms have achieved remarkable progress in low-light deblurring. However, the lack of large realistic training datasets is still an obstacle for DNN's generalization to practical applications. 

To handle the aforementioned problems systematically, we collect a new large dataset consisting of 3000 night images, and employ a more elaborate noise model that takes the physical pipeline of digital image acquisition into consideration to simulate realistic low-light blurry images. Besides, a novel physically-driven non-blind deblurring network incorporating Wiener deconvolution, DNN, and reblurring loss is proposed to tackle the challenges of blur, noise and saturation in low-light conditions. In the following sections, we will present the detailed design and provide extensive experimental validation.

\section{The Proposed Method}
\label{sec:our_method}

In this section, we present an image and feature space Wiener deconvolution network named INFWIDE to restore clear and sharp images from the low-light blurry ones. As illustrated in Fig.~\ref{fig:net_infwide}, INFWIDE has a two-branch architecture and can be divided into two parts:  image and feature space Wiener deconvolution, and multi-scale cross residual fusion. The former part employs the blurring model to serve as a physical prior and provides a coarse estimation of the sharp image from both image space and feature space. The latter part then incorporates a fusion network to fuse the intermediate results and generate the final deblurring output with the aid of learnable image prior. In the following, we will give a detailed introduction with respect to these two parts.

\subsection{Image and Feature Space Wiener Deconvolution}
Wiener deconvolution is a fast and widely used Fourier space non-blind deblurring algorithm based on the linear physical blurring model. As shown in Fig.~\ref{fig:net_infwide}, given a blurry image $\mathbf{y}$ and corresponding blur kernel $\mathbf{k}$, the deblurred image $\hat{\mathbf{x}}$ can be attained by Wiener deconvolution as
\begin{equation}
    G = \frac{\mathcal{F}(\mathbf{k})^*}{\mathcal{F}(\mathbf{k})^*\mathcal{F}(\mathbf{k}) + NSR}
\end{equation}

\begin{equation}
    \hat{\mathbf{x}} = \mathcal{F}^{-1}(G\odot\mathcal{F}(\mathbf{y})),
\end{equation}
where $\mathcal{F}$ and $\mathcal{F}^{-1}$ denote the discrete Fourier transform (DFT) and inverse DFT, respectively; $\mathcal{F}(\mathbf{k})^*$ is the complex conjugate of $\mathcal{F}(\mathbf{k})$; $\odot$ represents element-wise product. $NSR = \sigma_n^2/\sigma_s^2$ is the noise-to-signal ratio. In practice, $\sigma_s^2$ is estimated as the standard variation of the blurry input $\mathbf{y}$, and $\sigma_n^2$ is estimated as the standard variation of the difference between $\mathbf{y}$ and the mean-filtered result of $\mathbf{y}$.

Prior works have proved that feature space Wiener deconvolution has superior performance in suppressing artifacts and recovering fine details \cite{dong2020DeepWiener,dong2021DWDNDeep}. But in low-light conditions, the blurry images corrupted by complex noise and saturation have much less information, which imposes more challenge on subsequent feature extraction and deblurring process. Bearing this in mind, we employ a two-branch strategy in INFWIDE to deblur the low-light blurry image in both image space and feature space by investigating the complementary information. 

To deal with the saturation and noise problems, we first employ an enhancement module (EM) in the image branch of INFWIDE to restore a clear and non-clipped blurry image from the original input. Unlike existing works that estimate the saturation regions and directly discard them in subsequent processing, our method aims to recover a clear blurry image that obeys the linear blurring model shown in Eq.~\eqref{eq: blur_model}. In this manner, the subsequent Wiener deconvolution can work properly to generate a coarse deblurring result without loss of information, especially in saturation regions. The enhancement module has a ResUNet architecture \cite{zhang2020DeepUnfolding}, which replaces the convolutional layers with residual blocks in a conventional UNet. To improve the enhancement module's robustness to different noise levels, a noise level map is estimated and concatenated with the low-light blurry image to serve as the input. A similar noise level estimation method is applied as in $NSR$'s calculation.

In terms of feature space Wiener deconvolution, we first employ a feature module (FM) to extract a set of feature maps from the original low-light blurry input. Then, the Wiener deconvolution is conducted on the blurry feature maps to generate their sharp counterparts. And finally, a feature space refine module (FSRM) is employed to recover another coarse deblurring result from the deblurred feature maps. In our implementation, the feature module is composed of a convolutional layer and three residual blocks in a sequential manner, and the feature space refine module has a similar ResUnet architecture as the enhancement module.

\subsection{Multi-Scale Cross Residual Fusion}
After we have attained two intermediate deblurring results with the aid of physically-driven Wiener deconvolution in both image space and feature space, the next step is to fuse them together to generate the final sharp image with fine details. In order to make full use of the complementary information, a multi-scale cross residual fusion network is designed.

\begin{figure}[th]
  \centering
  \includegraphics[width=\linewidth]{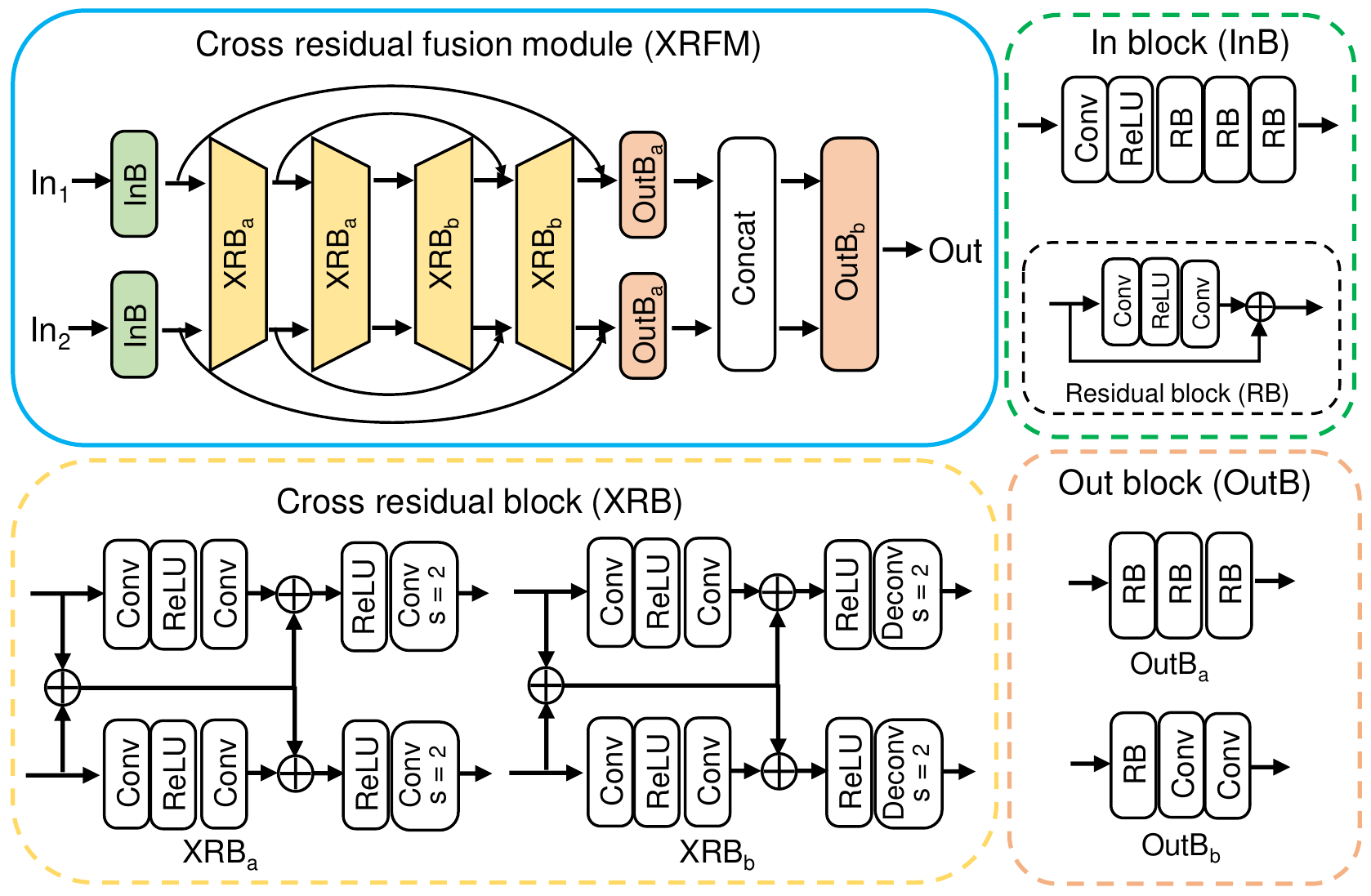}
  \caption{Detailed structure of the cross residual fusion module (XRFM).}
  \label{fig:net_xrfm}
\end{figure}

The cross residual fusion module (XRFM) features an encoder-decoder architecture, and mainly consists of four cross residual blocks (XRB) linked by skip connections as shown in Fig.~\ref{fig:net_xrfm}. The cross residual block is a generalized version of the residual block, which has two input channels and two output channels bridged by a shared inner residual connection. Compared with widely used "early-fusion" (i.e. concatenating/adding two branches at the beginning of the network) or "late-fusion" (i.e. concatenating/adding two branches at the end of the network), XRFM digs deeper into the use of complementary information from two branches by fusing them across multi-levels. Apart from the novel fusion module, a multi-scale refine strategy is also employed to progressively recover the fine details as many other deblurring approaches do \cite{zhang2022DeepImage}. To balance the deblurring quality and efficiency, we set the total scales of multi-scale cross residual fusion to two in our implementation, and XRFM in different scales shares the same network parameters during both training and testing.

\subsection{Loss Functions}
The loss function of INFWIDE mainly consists of three parts, i.e. deblurring loss $\mathcal{L}_{deblur}$, enhancement loss $\mathcal{L}_{enhance}$, and reblurring loss $\mathcal{L}_{reblur}$. As presented in Eq.~\eqref{eq:loss}, we empirically set the weights of these three parts to be 1, 0.5, and 0.5, although there might exist a better choice.
\begin{equation}
\label{eq:loss}
 \mathcal{L} = \mathcal{L}_{deblur} + 0.5\mathcal{L}_{enhance} + 0.5\mathcal{L}_{reblur}
\end{equation}

\vspace{1mm}
\noindent \textbf{Deblurring loss.} 
Deblurring loss $\mathcal{L}_{deblur}$ is the multi-scale loss between the deblurring results and corresponding ground truth images. It dominates the entire network's convergence direction, and is composed of $l_1$ loss, $l_2$ loss, TV loss, and structural  similarity index measure (SSIM) loss:

\begin{gather}
\label{eq:db_loss}
\mathcal{L}_{deblur} = \frac{1}{MN}\sum_{i=1}^N\Big(\sum_{l=1}^M\big(\gamma_1\|\mathbf{\hat{x}}_i^l - \mathbf{x}_i^l\|_1 + \gamma_2\|\mathbf{\hat{x}}_i^l - \mathbf{x}_i^l\|_2 \notag \\ 
+ \gamma_3\|\mathbf{\hat{x}}_i^l \|_{tv} + \gamma_4\mathcal{L}_{ssim}(\mathbf{\hat{x}}_i^l) \big)\Big).
\end{gather}
Here $\hat{\mathbf{x}}^l$ is the output of the cross residual fusion module at scale $l$; $\mathbf{x^l}$ represents the downsampled ground truth image using bicubic interpolation for scale $l$; $N$ is the batch size and $M$ is the number of scales in cross residual fusion. We empirically set the weights $\gamma_1$, $\gamma_2$, $\gamma_3$, and $\gamma_4$ to be 0.4, 0.2, 0.2, and 0.2 in our experiments.

\vspace{1mm}
\noindent \textbf{Enhancement loss.}
Enhancement loss $\mathcal{L}_{enhance}$ aims to guide the enhancement module in the image branch of INFWIDE to restore a sharp and non-clipped blurry image which obeys the linear convolution model shown in Eq.~\eqref{eq: blur_model}. In this manner, the subsequent Wiener deconvolution can efficiently generate a better intermediate deblurring result. To better regularize the enhancement module under sophisticated noise and saturation, $\mathcal{L}_{enhance}$ consists of three different loss types, i.e. $l_1$ loss, $l_2$ loss, and total variation (TV) loss as follows
\begin{equation}
\label{eq:eh_loss}
\begin{split}
\mathcal{L}_{enhance} = &\frac{1}{N}\sum_{i=1}^N \big(\gamma_1\|\mathbf{\hat{y}}_i - \mathbf{x}_i \ast \mathbf{k}_i\|_1 \\ &+ \gamma_2\|\mathbf{\hat{y}}_i - \mathbf{x}_i \ast \mathbf{k}_i\|_2 + \gamma_3\|\mathbf{\hat{y}}_i \|_{tv} \big),
\end{split}
\end{equation}
where $N$ is the batch size; $\mathbf{x}$, $\mathbf{k}$, and $\hat{\mathbf{y}}$ are the ground truth sharp image, blur kernel, and output of the enhancement module, respectively. The weights $\gamma_1$, $\gamma_2$, and $\gamma_3$ is empirically set to be 0.5, 0.3, and 0.2 in our experiments. 

\vspace{1mm}
\noindent \textbf{Reblurring loss.} 
Existing non-blind deblurring networks are generally trained in an end-to-end manner and employ only the deblurring loss between deblurring results and corresponding ground truth as their loss functions. However, they neglect the fact that non-blind deblurring is an inverse problem with an explicit analytical forward model, which can serve as a strong regularizer to guide the network's convergence towards the right direction. Therefore, in this work, we introduce a physics-constrained `reblurring loss' to introduce more physical prior to facilitate the training process:
\begin{equation}
\label{eq:rb_loss}
\begin{split}
\mathcal{L}_{reblur} = &\frac{1}{N}\sum_{i=1}^N\|\mathbf{\hat{x}}_i \ast \mathbf{k}_i - \mathbf{x}_i \ast \mathbf{k}_i\|_1
\end{split}
\end{equation}
Reblurring loss incorporates the forward blurring model (i.e. Eq.~\eqref{eq: blur_model}) to `reblur' the deblurring result $\hat{\mathbf{x}}$ of INFWIDE and minimizes the gap between the reblurred image and the corresponding ground-truth clear non-clipped blurry image. In this manner, we can efficiently build a close-loop regularization restrained by the physical prior to guarantee the network’s convergence within a feasible region. 

It is worth noting that the `reblurring loss' has also been used in some blind deblurring networks\cite{chen2018reblur2deblur,nah2021clean}. However, they have different underlying logic with the reblurring loss mentioned above. To be specific, they generally require an extra blur kernel estimation module \cite{chen2018reblur2deblur} or reblurring module \cite{nah2021clean} to simulate the `reblurring` operation for lack of ground truth blur kernels. Therefore, their reblurring losses don't introduce extra physics information for regularization as ours do, and thus should only be regarded as a self-constrained/self-consistent strategy.

\section{Low-light Dataset Simulation}
\label{sec:data_gen}

\subsection{Low-light Dataset Collection}
Dataset has a significant influence on the performance of learning-based algorithms. Existing low-light deblurring networks are generally trained on intensity-decayed daytime images deteriorated by Gaussian or Poisson noise. However, such images feature low HDR, uniform illumination and simplified noise, which cannot resemble real low-light scenes well. Chen et al. collected 600 nighttime images from Flickr to generate their dataset for low-light deblurring \cite{chen2021LearningNonBlind}. Considering that a large dataset is required for the training of deeper neural networks, we extend their dataset to 3200 images by collecting more images. 

Specifically, we collect low-light images from the photo-sharing websites like Flickr and Unsplash based on the following criteria: 1) The images should be taken in low-illumination scenarios, most of which should feature high dynamic range for the simulation of saturation regions.  2) The images should be well-exposed (generally with a longer exposure time due to the weak light condition) and contain little noise or blur. This is because we regard the collected images as ideal ground truth, and the corresponding inputs with noise and blur are synthesized by the physical-process-based noise model and blurring model. 3) The images should cover a variety of scenarios and objects, which helps to improve the trained model’s generalization ability. The new large dataset is named \textit{NightShot}. It's further divided into two separate parts as the training and testing datasets, which contain 3000 images and 200 images, respectively.

\subsection{Simulation of Blur, Noise, and Saturation}
\label{sec: simu_method}

\begin{figure}[t]
  \centering
  \includegraphics[width=\linewidth]{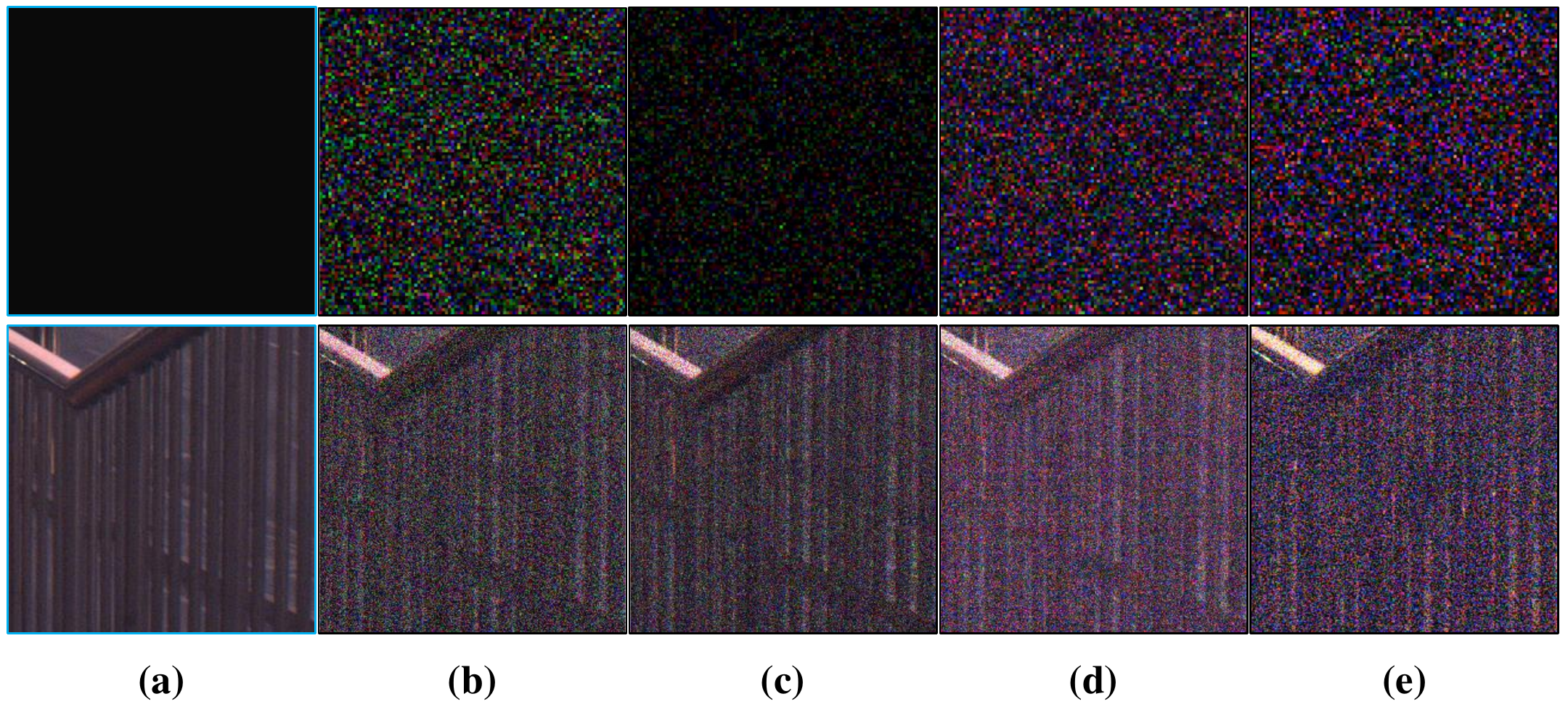}
  \caption{Comparison of noisy images synthesized by different noise models with the real-captured noisy images. (a) Well-exposed clean images for simulation. (b-d) Noisy images simulated by zero-mean Gaussian model, Poisson model, and our physical-process-based model, respectively. (e) Real-captured noisy images.}
  \label{fig:noise_simu}
\end{figure}

Images in \textit{NightShot} are generally well-exposed sharp images taken by experienced photographers. In order to generate realistic low-light blurring images from \textit{NightShot}, we synthesize blur, noise, and saturation successively according to the low-light blurring model in Eq.~\eqref{eq: night_blur_model}.

To be specific, we firstly generate random blur kernels ranging from $13 \times 13$ to $35 \times 35$ for each image in \textit{NightShot} with the code of \cite{schmidt2016CascadesRegression}. Then we convolve each image with the blur kernel to simulate its blurry counterpart. As mentioned in Sec.~\ref{sec:related_work_noise}, noise model in low-light conditions is more complex but plays a crucial role in learning-based deblurring algorithms' generalization in practical applications. Therefore, to synthesize realistic low-light noise, we employ a more comprehensive noise model \cite{cheng2022DualSensor, wang2019EnhancingLow}, which delicately takes the physical imaging process of camera sensors into consideration.

For a typical charge coupled device (CCD) or complementary metal-oxide-semiconductor (CMOS) image sensor, the digital image formation process can be roughly divided into three steps: \textit{i)} photon collection and photoelectric conversion; \textit{ii)} signal readout and  amplification; and \textit{iii)}  analog-to-digital conversion and post-processing. Noise with various features will be introduced during these steps. Generally, most works only take the dominant noise including shot noise and readout noise into consideration \cite{brooks2019unprocessing, miao2020HandlingNoise,corbineau2019LearnedImage}. However, in this work, considering that the dark current and dynamic streak phenomenon become non-negligible in low-light conditions and might influence the deblurring performance, we also involve them in our noise simulator apart from the shot noise and readout noise.

\vspace{1mm}
\noindent \textbf{Shot noise.} At the front-end of an imaging system, light from the scene is concentrated by lenses and arrives on pixel areas as photons. Due to the intrinsic stochasticity of photons reaching the sensor plane, inevitable signal-dependent shot noise will be generated and can be modeled with Poisson distribution. Specifically, the detected signal $S^i$ of the $i_{th}$ pixel contaminated by shot noise can be modeled as:
\begin{equation}
\label{eq:shot_noise}
S^i \sim \mathcal{P}(N^i_p)
\end{equation}
with $N^i_p$ denoting the photoelectron count in the $i_{th}$ pixel and $\mathcal{P}(\cdot)$ representing the Poisson distribution. 

\vspace{1mm}
\noindent \textbf{Dark current.} 
In photosensitive devices like CCD or CMOS, there is relatively small electric current caused by random generation of electrons and holes within the depletion region even when no photons are received \cite{konnik2014HighlevelNumerical}.  This signal-independent electric current is called dark current, and can be described using a clipped Poisson distribution \cite{cheng2022DualSensor}.  Assuming that the expected number of dark current electrons per pixel is $N_d$, noise $D^i$ caused by dark current can be expressed as
\begin{equation}
\label{eq:dark_cur}
D^i  = \max\{0, n_d - N_d\}, n_d \sim \mathcal{P}(N_d).
\end{equation}

\vspace{1mm}
\noindent \textbf{Readout noise.} 
Readout noise occurs when electrons are converted to voltage or current signals. This kind of noise is highly relevant to image sensors' readout rate, and obey the zero-mean Gaussian distribution. Mathematically, the readout noise can be written as
\begin{equation}
\label{eq:readout}
R^i  \sim \mathcal{G}(0, \sigma_r^2),
\end{equation}
where $\mathcal{G}(0,\cdot)$ denotes the zero-mean Gaussian ditribution, and $\sigma_r$ is the standard deviation of readout noise.

\vspace{1mm}
\noindent \textbf{Dynamic streak, quantization and gain.}
Apart from the aforementioned canonical noise, there are also some factors that can change the existing noise's distribution, thus resulting in different visual effects on final images. Dynamic streak is a common phenomenon in low-light imaging, which degrades the images with frame-varying horizontal streaks. In accordance with the fluctuation characteristic of the dynamic streak, we can model it with a row-wise gain $\beta_r \in  \mathcal{G}(1, \sigma_\beta)$ \cite{wang2019EnhancingLow}. The processes of quantization and gain in digital imaging also impose a significant influence on existing noise's distribution and final image visuality. Generally, the analog voltage/current signal will be amplified for $K_{a}$ times and then quantized to discrete gray values before a digital image can be generated. Besides, a digital gain  $K_{d}$  could also be applied to improve the digital image's brightness for a better visuality in post-processing, especially for low-light imaging. 

Taking all the noise and factors analyzed above into consideration, we can model the final gray value $y^i$ of the $i_{th}$ pixel as follows
\begin{equation}
\label{eq:all_noise}
y^i_c = K_{d,c}K_{a,c}\beta_{r,c}(S^i + D^i + R^i), c \in \{r,g,b\}.
\end{equation}
Note that due to the existence of color filter array (CFA) and non-uniform response of silicon devices for different color channels, the parameters of $K_d$, $K_a$, and $\beta_r$ vary with channels as shown above. Considering that $K_d$ and $K_a$ have similar functions, we simplify them with a total camera gain the variable $K = K_aK_d$ in our experiments in our experiments. 

In simulation, we assume that the original image $I$ from \textit{NightShot} is taken by an ideal imaging system under an illumination $E$ (i.e., totally free of noise and streak artifacts) and no amplification is introduced (i.e., camera gain $K=1$), so its intensities are proportional to the number of released photoelectrons $N_p$. Using $M$ to specify the brightness attenuation inversely proportional to the illumination level, we synthesize a realistic noisy low-light image taken under a weaker illumination $\frac{E}{M}$, by setting the photoelectron counts $N_p=\frac{I}{M}$ in our noise model. Besides, we also set the camera gain $K=M$ to make the synthesized image's brightness similar to $I$ to avoid further image enhancement after deblurring. 

Fig.~\ref{fig:noise_simu} demonstrates our synthetic noisy images and their comparison with images by other simulation methods and the real-captured noisy images. We can clearly find that our noise model produces more realistic simulation than other methods.

Finally, in terms of saturation regions' synthesis, we follow prior works' approach, i.e. enlarging the gray values of the blurry noisy images by a factor of 1.2 and clipping them back to the normal dynamic range of 0 to 1 afterwards \cite{ren2018DeepNonBlind, chen2021LearningNonBlind}. In this manner, pixels in some bright areas will saturate and violate the linear blurring model, which resembles the low-light imaging scenario well as expected.

\section{Experiments and Discussion}
\label{sec:exp}
\subsection{Implementation Details}
The training set is generated by successively synthesizing blur, noise, and saturation on \textit{NightShot} dataset according to the method described in Sec.~\ref{sec: simu_method}. To improve the network's robustness to different noise levels, we randomly sample the noise parameters within a range as listed in TABLE~\ref{tab:noise_param}.

\begin{table}[t]
  \renewcommand{\arraystretch}{1.8}
  \small
  \centering
  \vspace{1mm}
\caption{Parameter settings of the physical-process-based noise model for training set generation. Each parameter is randomly sampled from a uniform distribution to increase the data diversity.}
\begin{tabular}{c|c c c c}
    \hline
    \textbf{Parameters} & $N_d$ & $\sigma_{r}$ & $\sigma_\beta$ & $K$ \\ \hline
    \textbf{Distribution} & $U(2, 8)$ & $U(0.5,4)$ & $U(0.01,0.03)$ & $U(4,16)$\\ \hline
  \end{tabular}
  \label{tab:noise_param}
\end{table}

We implement the proposed network with PyTorch\footnote{The code will be available at https://github.com/zhihongz/INFWIDE}. To update the network's parameters, an Adam optimizer is adopted by setting $\beta_1 = 0.9$, $\beta_2 = 0.999$, and $\epsilon=1e^{-8}$. The initial learning rate is $2e^{-4}$, and it decays by a factor of 0.5 for every 50 epochs. The training process is conducted on randomly cropped image patches with the size of $256 \times 256$ pixels, and the batch size is set to 8. All experiments are performed on a workstation equipped with an NVIDIA GeForce RTX 3090 GPU and an AMD Ryzen Threadripper 3970X 32-Core CPU with 256G memory.

To prove the superior performance of INFWIDE on low-light non-blind deblurring, we compare it with state-of-the-art approaches including Cho et al. \cite{cho2011HandlingOutliers}, Hu et al.\cite{hu2014DeblurringLowlight}, Whyte et al. \cite{whyte2014DeblurringShaken}, Sanghvi et al.\cite{sanghvi2021PhotonLimited}, RGDN \cite{gong2020LearningDeep}, IRCNN \cite{zhang2017LearningDeep}, and DWDN \cite{dong2020DeepWiener} on both synthetic and real data\footnote{We didn't compare with NBDN \cite{chen2021LearningNonBlind} as we couldn't find or reproduce its training codes.}. The first three competing methods are optimization-based and the rest are learning-based. For a fair comparison, we re-train all the learning-based methods on our training set. Besides, we use the camera gain $K$ to serve as a substitution for the photon level $\alpha$ in Sanghvi's method, as both variables represent the light flux. For IRCNN, a noise level estimation is required beforehand, so we try different noise levels and select the most proper one. It is worth noting that INFWIDE doesn't require information of light-level or noise-level during both training and testing, which makes it more flexible and universal in practical applications.

\subsection{Results with Synthetic Data}

\begin{table*}[bht]
  \renewcommand{\arraystretch}{1.8}
  \small
  \centering
  \vspace{1mm}
\caption{Performance comparison of the proposed method (INFWIDE) with other competing approaches on \textit{NightShot} test dataset in terms of average  PSNR(dB) / SSIM}
\resizebox{1.0\textwidth}{!}{
\begin{tabular}{c|c c c c c c c c}
    \hline
    Camera gain & Cho \cite{cho2011HandlingOutliers} & Hu \cite{hu2014DeblurringLowlight} & Whyte \cite{whyte2014DeblurringShaken} & Sanghvi \cite{sanghvi2021PhotonLimited} & RGDN \cite{gong2020LearningDeep} & IRCNN \cite{zhang2017LearningDeep} & DWDN \cite{dong2020DeepWiener} & INFWIDE (ours) \\ 
    \hline
    K = 4 & 22.89 / 0.8129 & 20.88 / 0.6498 & 11.52 / 0.2791 & 25.14 / 0.7650 & 19.48 / 0.5397 & 25.50 / 0.7626 & 25.74 / 0.7828 & \textbf{26.51 / 0.8130}\\ 
    \hline
    K = 8 & 21.80 / 0.7555 & 18.22 / 0.5109 & 9.14 / 0.1871 & 24.51 / 0.7375 & 19.49 / 0.5363 & 24.70 / 0.7169 & 24.82 / 0.7511 & \textbf{25.45 / 0.7798}\\ 
    \hline
    K = 16 & 19.97 / 0.6724 & 13.87 / 0.3524 & 6.24 / 0.1020 & 23.31 / 0.6855 & 19.30 / 0.5101 & 22.78 / 0.6042 & 23.60 / 0.7029 & \textbf{24.14 / 0.7329}\\ 
    \hline
  \end{tabular}
}
\label{tab:simu_exp_ns}
\end{table*}

\begin{table*}[bht]
  \renewcommand{\arraystretch}{1.8}
  \small
  \centering
  \vspace{1mm}
\caption{Performance comparison of the proposed method (INFWIDE) with other competing approaches on test dataset from Hu et al. \cite{hu2014DeblurringLowlight} in terms of average  PSNR(dB) / SSIM}
\resizebox{1.0\textwidth}{!}{
\begin{tabular}{c|c c c c c c c c}
    \hline
    Camera gain & Cho \cite{cho2011HandlingOutliers} & Hu \cite{hu2014DeblurringLowlight} & Whyte \cite{whyte2014DeblurringShaken} & Sanghvi \cite{sanghvi2021PhotonLimited} & RGDN \cite{gong2020LearningDeep} & IRCNN \cite{zhang2017LearningDeep} & DWDN \cite{dong2020DeepWiener} & INFWIDE (ours) \\ 
    \hline
    K = 4 & 26.66 / 0.7468 & 22.58 / 0.4047 & 16.41 / 0.1483 & 29.29 / 0.8494 & 23.34 / 0.6609 & 29.35 / 0.8291 & 28.78 / 0.8334 & \textbf{29.76 / 0.8703}\\ 
    \hline
    K = 8 & 25.14 / 0.6401 & 18.83 / 0.2261 & 12.95 / 0.0732 & 28.54 / 0.8194 & 23.30 / 0.6496 & 28.53 / 0.7564 & 27.89 / 0.8085 & \textbf{28.67 / 0.8435}\\ 
    \hline
    K = 16 & 22.87 / 0.5719 & 14.31 / 0.1146 & 8.98 / 0.0302 & 27.37 / 0.7743 & 23.01 / 0.6103 & 26.49 / 0.6301 & 26.95 / 0.7725 & \textbf{27.57 / 0.8106}\\ 
    \hline
  \end{tabular}
}
\label{tab:simu_exp_hu}
\end{table*}

\begin{figure*}[t]
  \centering
  \includegraphics[width=\linewidth]{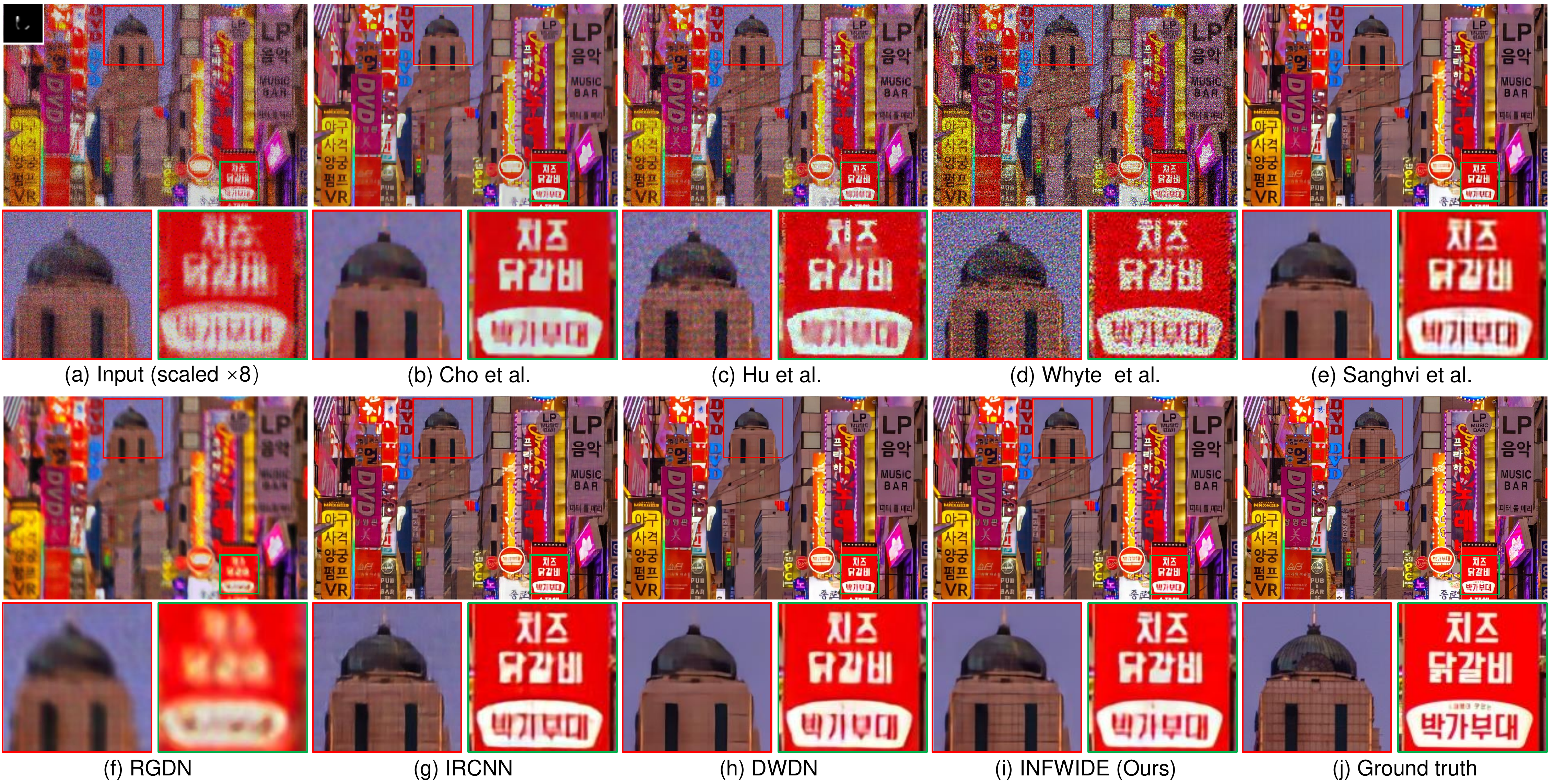}
  \caption{Qualitative comparison of INFWIDE with other state-of-the-art non-blind deblurring algorithms \cite{cho2011HandlingOutliers, hu2014DeblurringLowlight,whyte2014DeblurringShaken,sanghvi2021PhotonLimited, gong2020LearningDeep, zhang2017LearningDeep, dong2020DeepWiener} on a low-light blurry image from \textit{NightShot} dataset ($K$=8), with severe noise and saturation. 
  (please zoom-in for a better view).}
  \label{fig:simu_exp_ns}
\end{figure*}

We first evaluate INFWIDE and the competing methods quantitatively on two synthetic datasets, i.e. the test set of \textit{NightShot} dataset and a popular low-light deblurring benchmark from Hu et al. \cite{hu2014DeblurringLowlight}. The \textit{NightShot} test set contains 200 images. We synthesize 100 random blur kernels and convolve each blur kernel with two different images generating 200 blurry images for testing. The dataset from Hu et al. consists of 14 night images and 11 blur kernels. We convolve each blur kernel with all the images and get 154 blurry images in total. Note that noise and saturation are also simulated according to the methods described in Sec.~\ref{sec: simu_method}, and we examine the deblurring performance under different noise levels separately by controlling the camera gain.

The average peak signal to noise ratio (PSNR) and structural similarity index measure (SSIM) for different deblurring methods are summarized in Table~\ref{tab:simu_exp_ns} and Table~\ref{tab:simu_exp_hu}. As can be seen from the tables, INFWIDE outperforms all other approaches in terms of PSNR and SSIM under all noise levels on both \textit{NightShot} dataset and Hu et al's dataset. In comparison, the learning-based algorithms \cite{sanghvi2021PhotonLimited, dong2020DeepWiener,chen2021LearningNonBlind} generally achieve superior performance to optimization-based algorithms \cite{cho2011HandlingOutliers,hu2014DeblurringLowlight,whyte2014DeblurringShaken} by digging deeper into data-driven image priors. Although RGDN \cite{dong2020DeepWiener} also employs learnable modules in its optimization framework, it depends heavily on the linear blurring model and has little ad-hoc design for saturation and sophisticate noise problems, thus failing to attain similar performance on par with other learning-based algorithms. 

\begin{figure*}[t]
  \centering
  \includegraphics[width=\linewidth]{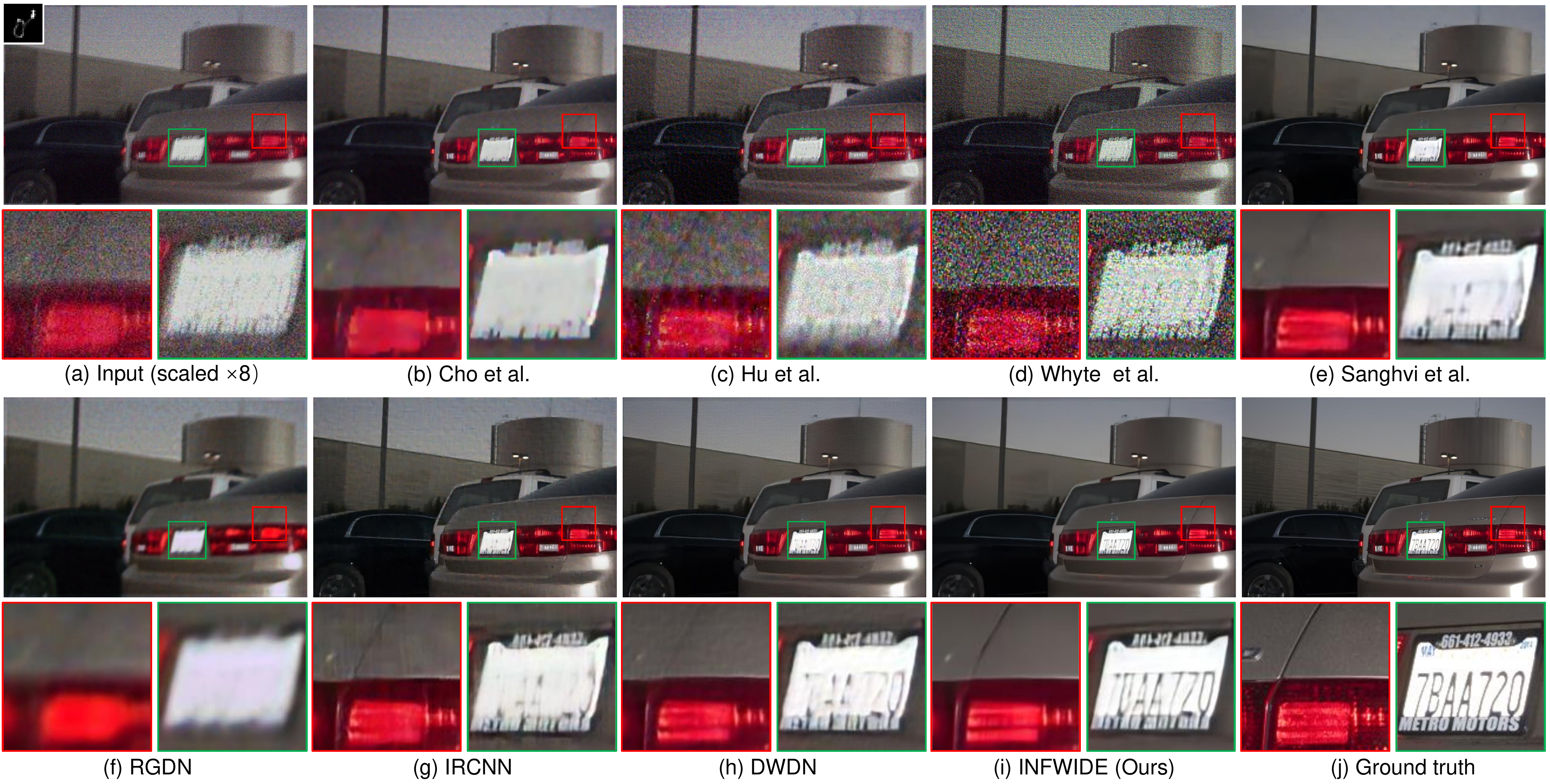}
  \caption{Qualitative comparison of INFWIDE with other state-of-the-art non-blind deblurring algorithms on a low-light blurry image from Hu et al.'s dataset \cite{hu2014DeblurringLowlight} (K=8). 
  Please zoom-in for a better view.}
  \label{fig:simu_exp_hu}
\end{figure*}

\label{sec:real_exp}
\begin{figure*}
  \centering
  \includegraphics[width=\linewidth]{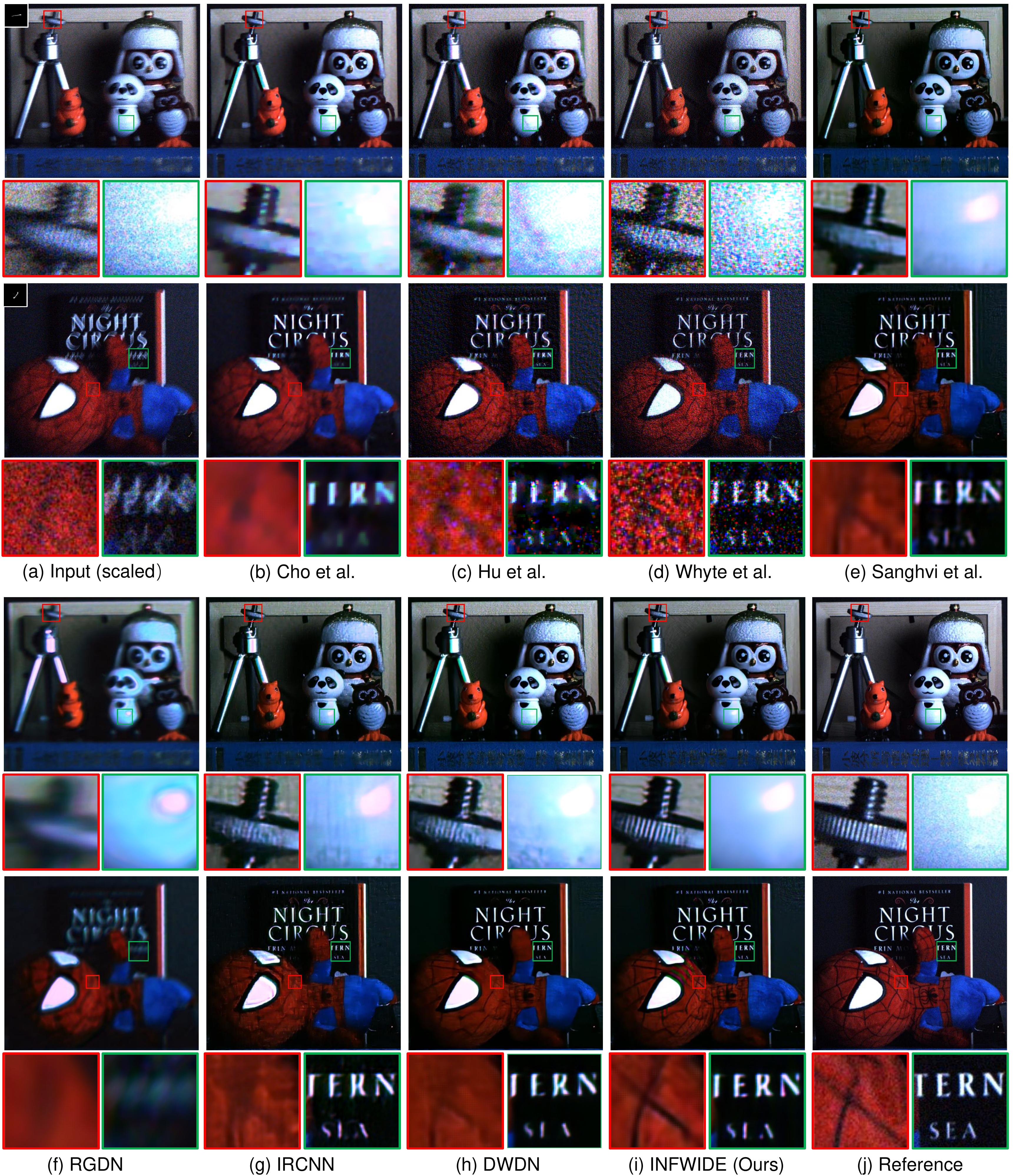}
  \caption{Comparison of INFWIDE with other non-blind deblurring algorithms on a real-captured low-light image (5 lux) from Sanghvi et al. \cite{sanghvi2021PhotonLimited}. 
  Please zoom-in for a better view.}
  \label{fig:real_exp}
\end{figure*}

We further demonstrate some low-light deblurring examples from INFWIDE and competing approaches for visual comparison in Fig.~\ref{fig:simu_exp_ns} and Fig.~\ref{fig:simu_exp_hu}. As can be seen from the figures, due to the existence of complicated noise and saturation problem in low-light conditions, optimization-based methods suffers severe artifacts in the deblurred images, while learning-based algorithms tend to over-smoothen the deblurring results. On the contrary, INFWIDE can recover the fine details and prevent unpleasant artifacts during deblurring.

\subsection{Results on Real Data}

To demonstrate INFWIDE's superior performance in practical applications, we qualitatively compare INFWIDE and other competing methods on a real-captured low-light image blurred by camera shake from \cite{sanghvi2021PhotonLimited}. The image is taken using a Canon EOS Rebel T6i camera under 5 lux illumination, and the blur kernel is calibrated with the aid of a point source placed in the scene. Before deblurring, symmetric padding is conducted to mitigate the influence of the non-circular convolution issues in real-captured blurry images.

As shown in Fig.~\ref{fig:real_exp}, the low-light blurry image contains severe noise and saturation regions caused by highlights. Nevertheless, INFWIDE can still restore the latent clear and sharp image efficiently. The deblurring result of INFWIDE features more fine textures and fewer artifacts compared with other methods. Even around the saturation regions, INFWIDE can still work well to recover the delicate gradual change of the highlight. This experiment reveals the excellent generalization ability of INFWIDE trained on our physical-process-based synthetic data in practical scenarios.

\begin{figure}[ht]
  \centering
  \includegraphics[width=\linewidth]{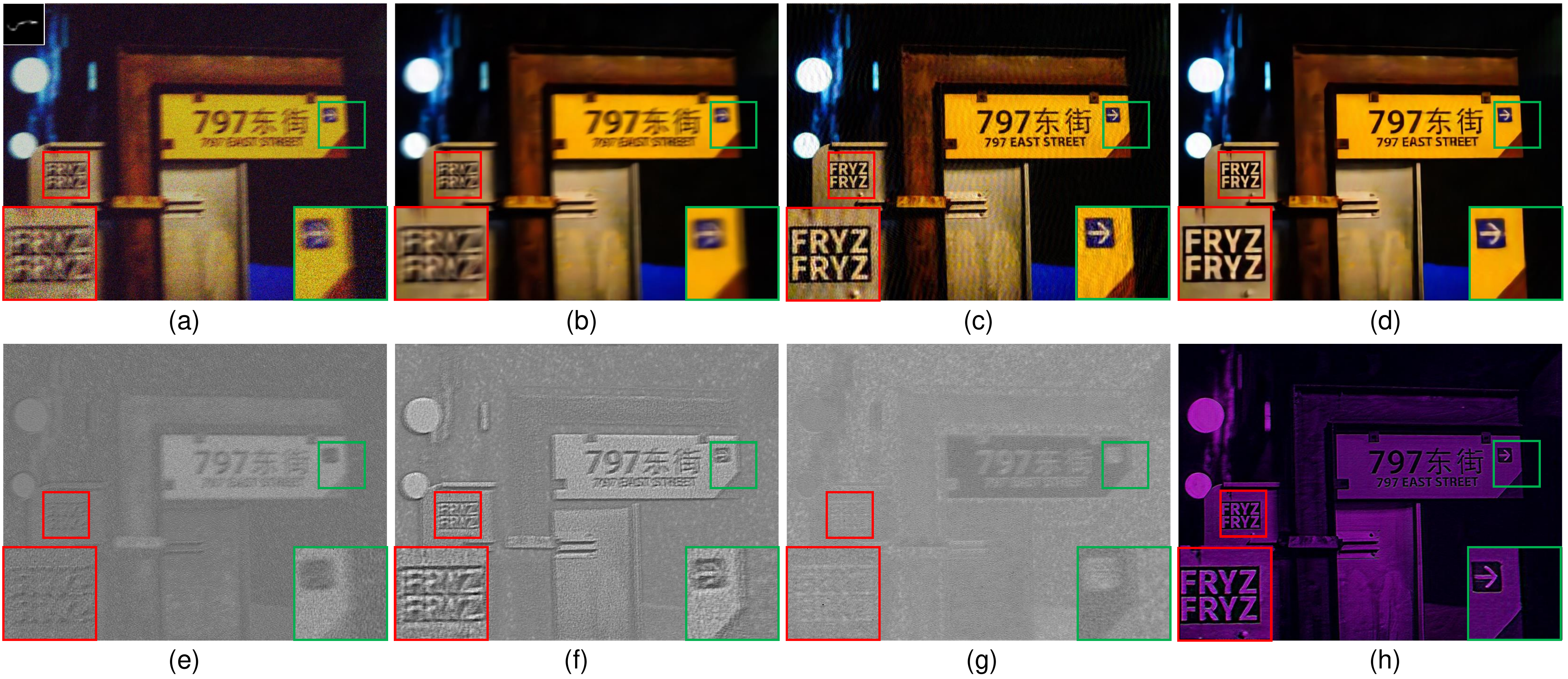}
  \caption{Visualization of the intermediate results. (a) Low-light blurry input. (b) Enhancement module's output. (c) Image branch's output. (d) Final deblurring result. (e-g) Feature module's outputs (3 feature maps out of 16 are shown). (h) Feature branch's output.}
  \label{fig:feature_analysis}
\end{figure}

\subsection{Ablation Studies and Analysis}
\subsubsection{Two-branch Network Architecture}
In the designed network architecture, we employ a two-branch fusion strategy to exploit the complementary information in both image space and feature space. Moreover, an enhancement module is incorporated in the image branch to recover a clear non-clipped image, and a feature module is incorporated in the feature branch to extract the useful features. In order to investigate what these modules and branches actually learn after training, we visualize their outputs in Fig.~\ref{fig:feature_analysis}. As can be seen from Fig.~\ref{fig:feature_analysis}(b), the enhancement module efficiently removes the severe noise and compensates for the clipping effect occurring in the low-light blurry input as expected. In contrast, the outputs of the feature module (Fig.~\ref{fig:feature_analysis}(e-g)) contain various structural information but are still contaminated by obvious noise, which suggests that the feature module mainly plays the role of extracting useful features rather than denoising the input as the enhancement module does. 

Based on these two modules, the image branch and the feature branch could focus on different aspects and retrieve complementary information for deblurring. As shown in Fig.~\ref{fig:feature_analysis}(c) and (h), the output of the image branch features color fidelity and high dynamic range, but suffers from some ripple artifacts. Conversely, the feature branch’s output brings about clean regions and sharp structures while involving large deviation of color tones and brightness. By fusing the outputs of these two branches with the designed cross residual fusion module, our method can combine their strength and achieve a significant promotion in the final deblurring result (Fig.~\ref{fig:feature_analysis}(d)).

To quantitatively demonstrate the advantage of the two-branch architecture design, we further compare INFWIDE with its single-branch (i.e. only image branch or only feature branch) counterparts on \textit{NightShot} dataset. The results are summarized in Table~\ref{tab:ablation_arch}. INFWIDE-I and INFWIDE-F in the table represent the architectures with a single image branch or a single feature branch, respectively. It can be found that the proposed two-branch architecture INFWIDE achieves a significant improvement compared with the single-branch architectures. This demonstrates the strength of INFWIDE in exploiting the complementary information of both image space deconvolution and feature space deconvolution results.

\subsubsection{Multi-scale Cross Residual Fusion}
The design of the two-branch fusion module plays a significant role in INFWIDE. As the image branch and the feature branch focus on different aspects in the deblurring process, their intermediate outputs have significantly different characteristics, which makes it hard for `shallow-fusion' strategies like widely used addition or concatenation to fuse such two branches efficiently.

By contrast, the proposed multi-scale cross residual fusion module (XRFM) can provide a deeper and finer fusion effect. On the one hand, it builds on the specially designed cross residual block (XRB) which treats two branches differently by extracting their features with two separate Conv-layer streams instead of simply adding or concatenating them across the channel dimension. Meanwhile, XRB also introduces a cross residual path to link these two streams and dig out their complementary information efficiently. On the other hand, the proposed fusion module also employs the multi-scale strategy originating from two perspectives, i.e., the multiple levels in the basic U-Net structure of XRFM and the recursive multi-scale inputs in the coarse-to-fine fusion stage. By fusing two branches across multiple finer scales, INFWIDE can make more adequate use of the complementary information to improve the restored image’s quality.

To quantitatively validate the effectiveness of the cross residual fusion module, we also conduct another ablation experiment by replacing it with the frequently-used feature fusion modules, i.e., addition+ResUNet and concatenation+ResUNet. As shown in Table~\ref{tab:ablation_arch}, INFWIDE attains constantly superior performance to the addition-based (INFWIDE-A) and concatenation-based (INFWIDE-C) architectures over all noise levels (controlled by the camera gain K), and the specific average PSNR gains are 0.22dB and 0.10dB, respectively.

\begin{table}[ht]
  \renewcommand{\arraystretch}{1.8}
  \small
  \centering
  \vspace{1mm}
\caption{Ablation study of INFWIDE's architecture in terms of average  PSNR(dB) / SSIM on \textit{NightShot} dataset. INFWIDE-I and INFWIDE-F represent the architectures with only image branch and only feature branch, respectively. INFWIDE-A and INFWIDE-C represent the architectures with addition or concatenation fusion strategy instead of our XRFM.}
\resizebox{0.48\textwidth}{!}{
\begin{tabular}{c|c c c c c}
    \hline
    Camera gain & INFWIDE-I & INFWIDE-F & INFWIDE-A & INFWIDE-C & INFWIDE \\ 
    \hline
    K = 4 & 25.84 / 0.7925 & 26.16 / 0.8044 & 26.28 / 0.8079 & 26.41 / 0.8118  & \textbf{26.51 / 0.8130}\\ 
    \hline
    K = 8 & 24.89 / 0.7592 & 25.16 / 0.7725 & 25.26 / 0.7764 & 25.36 / 0.7786 & \textbf{25.45 / 0.7798}\\ 
    \hline
    K = 16 & 23.60 / 0.7105 &  23.92 / 0.7263 & 23.91 / 0.7297 & 24.03 / 0.7320 & \textbf{24.14 / 0.7329}\\ 
    \hline
  \end{tabular}
}
\label{tab:ablation_arch}
\end{table}


\subsubsection{Reblurring Loss}
Apart from the physical prior contained in the Wiener deconvolution based network architecture, we also employ an extra physical constraint by introducing a reblurring loss during training. The reblurring loss aims to guarantee that the deblurred image can be blurred back to the original clear and non-clipped blurry image, so that a close-loop regularization can be built to guide the convergence towards a more reasonable direction.

We conduct an ablation experiment on both INFWIDE and DWDN \cite{dong2020DeepWiener} to demonstrate the reblurring loss's architecture-independent effectiveness and summarize the results in Table~\ref{tab:ablation_reblur}. 
It can be observed that, by introducing the reblurring loss, INFWIDE attains an average improvement of 0.23dB in PSNR and 0.0023 in SSIM, while DWDN attains an average improvement of 0.13dB in PSNR and 0.0027 in SSIM.

\begin{table}[ht]
  \renewcommand{\arraystretch}{1.8}
  \small
  \centering
  \vspace{1mm}
\caption{Ablation study of the contribution of the reblurring loss in terms of average  PSNR(dB) / SSIM on \textit{NightShot} dataset.}
\resizebox{0.48\textwidth}{!}{
\begin{tabular}{c|c|c|c|c}
    \hline
    \multirow{2}{*}{Camera Gain}&
    \multicolumn{2}{c|}{INFWIDE}&
    \multicolumn{2}{c}{DWDN} \\
    \cline{2-5}
    & w/o $\mathcal{L}_{reblur}$ & w/ $\mathcal{L}_{reblur}$ & w/o $\mathcal{L}_{reblur}$ & w/ $\mathcal{L}_{reblur}$\\
    \hline
    K = 4 & 26.25 / 0.8101  & 26.51 / 0.8130 & 25.74 / 0.7828 & 25.93 / 0.7866  \\ 
    \hline
    K = 8 & 25.22 / 0.7774 & 25.45 / 0.7798 & 24.82 / 0.7511 & 24.94 / 0.7542 \\ 
    \hline
    K = 16 & 23.93 / 0.7312 & 24.14 / 0.7329 & 23.60 / 0.7029 & 23.69 / 0.7040 \\ 
    \hline
\end{tabular}

}
\label{tab:ablation_reblur}
\end{table}

\subsubsection{Physical-Process-Based Noise Model}
As mentioned before, the training data plays a significant role in deblurring networks' generalization to practical applications. Therefore, in this paper, we collect a large low-light dataset and employ a comprehensive physical-process-based noise model to synthesize realistic low-light blurry images for training. 

\begin{figure}[ht]
  \centering
  \includegraphics[width=\linewidth]{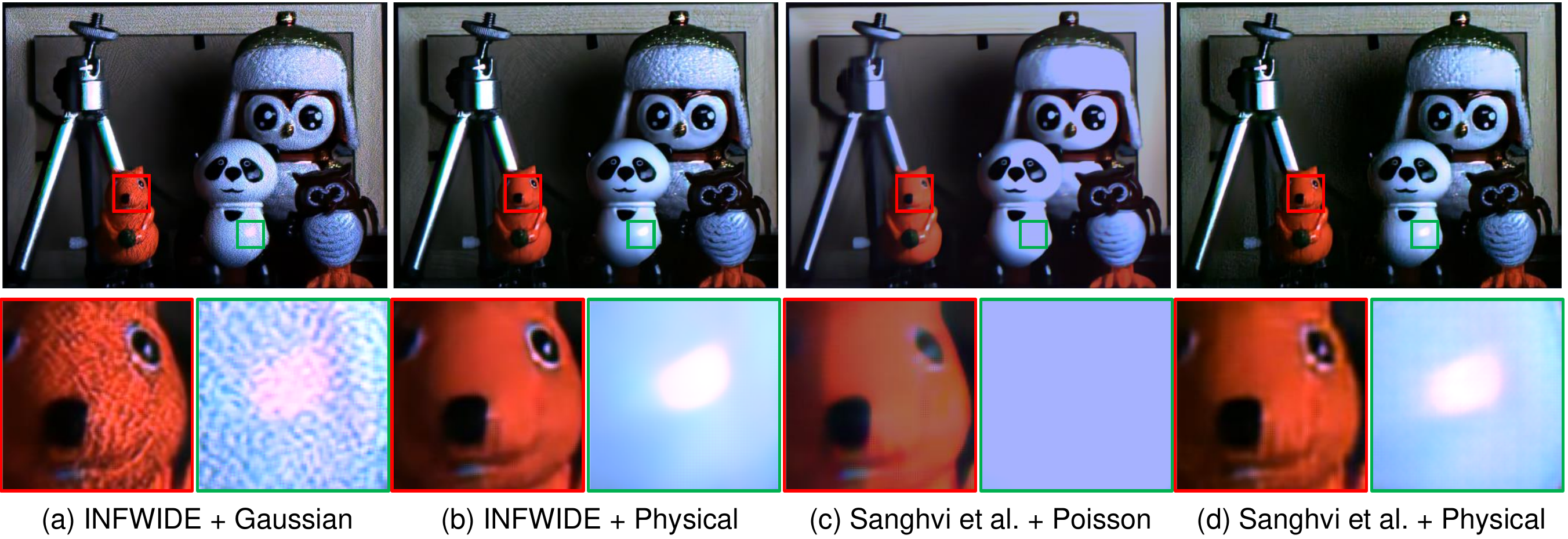}
  \caption{Influence of noise model  on real-data deblurring performance. (a) INFWIDE trained using Gaussian noise model. (b) INFWIDE trained using physical-process-based noise model. (c) Sanghvi's network trained using Poisson noise model. (d) Sanghvi's network trained using physical-process-based noise model.}
  \label{fig:noise_model_compare}
\end{figure}

In order to validate the contribution of the physical-process-based noise model in low-light deblurring, we train the INFWIDE using synthetic data corrupted by Gaussian noise as most existing works do, and compare its real-data deblurring result with the network trained using the physical-process-based noise model. As can be seen from Fig.~\ref{fig:noise_model_compare}(a) and (b), on account of the noise model mismatch between the synthetic data and real data, the deblurred image of INFWIDE trained using Gaussian noise model suffers severe artifacts. On the contrary, the network trained using the physical-process-based noise model have much fewer artifacts. We further demonstrate the advantage of the physical-process-based noise model versus Poisson noise model in real-data low-light deblurring using another network from Sanghvi et al. \cite{sanghvi2021PhotonLimited}. Note that Sanghvi's network is specially designed for coping with Poisson noise in photon-limited scenarios, and achieves state-of-the-art performance on synthetic datasets. As can be seen from Fig.~\ref{fig:noise_model_compare}(c) and (d), the network trained using Poisson noise is prone to be over-smoothing, especially around the saturation regions. However, this issue can be efficiently tackled by simply employing the physical-process-based noise model rather than Poisson noise model. 

\subsubsection{Parameter and Multiply–Accumulate Operation (MAC) Comparison}
The parameter and MAC numbers of different learning-based methods are summarized in Table~\ref{tab:param_macs}, in which the MAC numbers are calculated using ptflops\footnote{https://github.com/sovrasov/flops-counter.pytorch} with 256 × 256-pixel input. As can be seen from the table, INFWIDE has moderate Params/MACs numbers among these methods. Note that even though IRCNN \cite{zhang2017LearningDeep} and RGDN \cite{gong2020LearningDeep} have much fewer parameters, their MAC numbers are not proportionably small due to their multi-stage unrolling network architectures. Jointly considering the Params/MACs with performance, we could also find that although networks with more parameters tend to have stronger representation ability, they don’t necessarily get better performance. For example, while IRCNN \cite{zhang2017LearningDeep} has the fewest parameters, its overall performance exceeds RGDN \cite{gong2020LearningDeep} and is even on par with Sanghvi \cite{sanghvi2021PhotonLimited} and DWDN \cite{dong2020DeepWiener}. These observations suggest that the architecture design largely determines the final performance, and as validated by the ablation study our two-branch fusion architecture is effective in low-light deblurring.

\begin{table}[t]
  \renewcommand{\arraystretch}{1.8}
  \small
  \centering
  \vspace{1mm}
\caption{Parameter and MAC comparison for learning-based methods.}
\resizebox{0.48\textwidth}{!}{
\begin{tabular}{c|c c c c c}
    \hline
    \textbf{Models} & IRCNN \cite{zhang2017LearningDeep} & RGDN \cite{gong2020LearningDeep} &  DWDN \cite{dong2020DeepWiener} & INFWIDE & Sanghvi \cite{sanghvi2021PhotonLimited} \\ \hline
    \textbf{Params (M)} & 0.19 & 1.26 & 7.05 & 15.74 & 17.09 \\ \hline
    \textbf{MACs (G)} & 98.85 & 412.91 & 86.95 & 218.21 & 607.65 \\ \hline
  \end{tabular}
  }
  \label{tab:param_macs}
\end{table}

\section{Conclusion}
\label{sec:convlusion}
In this paper, we propose a novel physically-driven non-blind deblurring network named INFWIDE to cope with the sophisticated noise and saturation problem in low-light conditions. INFWIDE takes advantage of both image space and feature space Wiener deconvolution in a two-branch architecture to recover the fine details from different perspectives, and then employs a multi-level multi-scale fusion module to fuse both branches and generate the final clear and sharp image. To dig deeper into the physical prior contained in the blurring model, we further introduce a reblurring loss to serve as a close-loop regularization during training. Besides, we also collect a large low-light dataset and employ a comprehensive physical-process-based noise model to improve INFWIDE's generalization ability on real low-light blurry images.  

So far, INFWIDE can only deal with blurry images degraded by uniform blur kernels, which hinders its application in more complex scenarios. Patch-wise processing could be a simple but direct way to extend INFWIDE to non-uniform deblurring, but developing a more elegant and effective approach is still a meaningful research direction in the future. Besides, considering that the performance of INFWIDE is affected by the accuracy of blur kernel estimation in practical applications, thus improving INFWIDE's robustness to inaccurate blur kernels or developing more accurate blur kernel estimation algorithms for low-light blurry images is also a promising direction to explore in future work.

\section*{Acknowledgments}
This work was supported by the Ministry of Science and Technology of the People’s Republic of China [grant number 2020AAA0108202] and the National Natural Science Foundation of China [grant numbers 61931012, 62088102].



\bibliographystyle{IEEEtran}
\bibliography{paper_bib}

\vspace{11pt}
\begin{IEEEbiography}[{\includegraphics[width=1in,height=1.25in,clip,keepaspectratio]{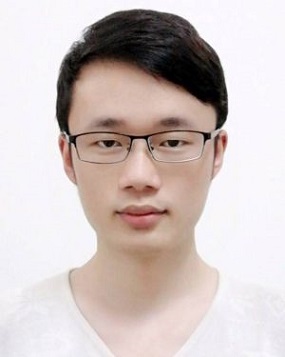}}]{Zhihong Zhang}
received the B.Eng. degree from the School of Electronics Engineering, Xidian University, Xi'an, China, in 2019. He is currently pursuing the Ph.D. degree in the Department of Automation, Tsinghua University. His research interests include computational imaging, computer vision, and machine learning.
\end{IEEEbiography}

\begin{IEEEbiography}[{\includegraphics[width=1in,height=1.25in,clip,keepaspectratio]{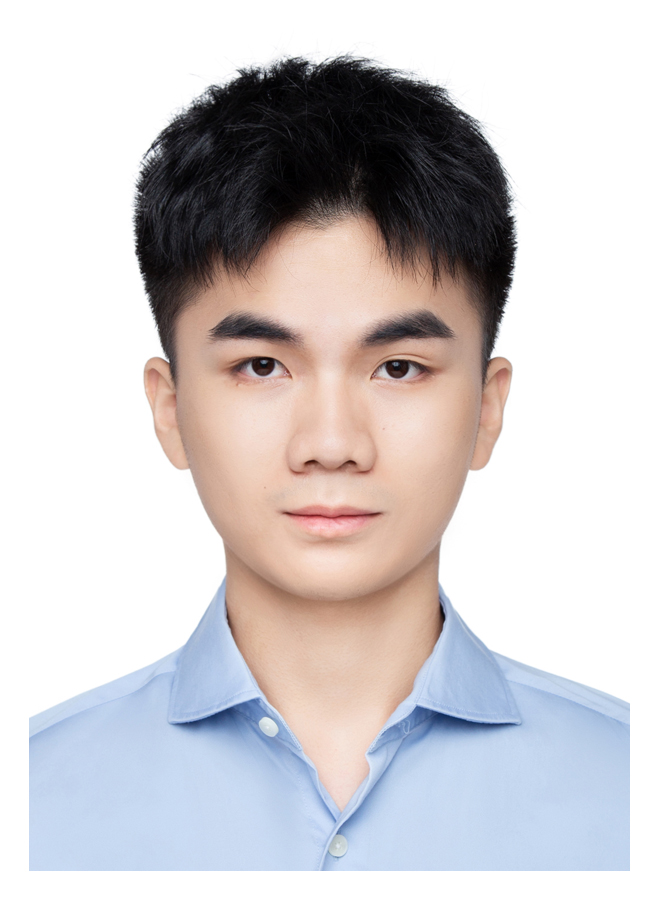}}]{Yuxiao Cheng}
received the B.Eng. degree from the Department of Automation, Tsinghua University, Beijing, China, in 2022. He is currently a Ph.D. student in the Department of Automation, Tsinghua University. His research interests include computational imaging and computer vision.
\end{IEEEbiography}

\begin{IEEEbiography}[{\includegraphics[width=1in,height=1.25in,clip,keepaspectratio]{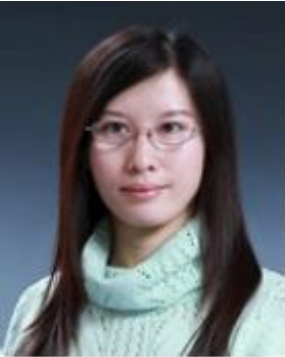}}]{Jinli Suo}
received the B.S. degree in computer science from Shandong University, Shandong, China, in 2004 and the Ph.D. degree from the Graduate University of Chinese Academy of Sciences, Beijing, China, in 2010. She is currently an Associate Professor with the Department of Automation, Tsinghua University, Beijing. Her current research interests include computer vision, computational photography, and statistical learning.
\end{IEEEbiography}

\begin{IEEEbiography}[{\includegraphics[width=1in,height=1.25in,clip,keepaspectratio]{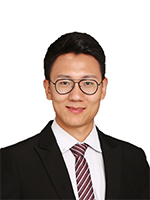}}]{Liheng Bian}
received the Ph.D. degree from the Department of Automation, Tsinghua University, Beijing, China, in 2018. He is currently an Associate Research Fellow with Beijing Institute of Technology. His research interests include computational imaging and computational sensing. More information can be found at https://bianlab.github.io/.
\end{IEEEbiography}

\begin{IEEEbiography}[{\includegraphics[width=1in,height=1.25in,clip,keepaspectratio]{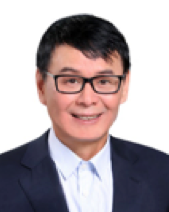}}]{Qionghai Dai}
(SM’05) received the ME and PhD degrees in computer science and automation from Northeastern University, Shenyang, China, in 1994 and 1996, respectively. He has been the faculty member of Tsinghua University since 1997. He is currently a professor with the Department of Automation, Tsinghua University, Beijing,
China, and an adjunct professor in the School of Life Science, Tsinghua University. His research areas
include computational photography and microscopy, computer vision and graphics, brain science, and video
communication. He is an associate editor of the JVCI, the IEEE TNNLS, and the IEEE TIP. He is an academician of the Chinese Academy of Engineering. He is a senior member of the IEEE.
\end{IEEEbiography}

\vfill

\end{document}